\pdfoutput=1
\documentclass[11pt]{article}

\usepackage[preprint]{acl}

\usepackage{times}
\usepackage{latexsym}
\usepackage[T1]{fontenc}
\usepackage[utf8]{inputenc}
\usepackage{microtype}
\usepackage{inconsolata}
\usepackage{amsthm}
\usepackage{amsmath}
\usepackage{amssymb}
\usepackage{mathrsfs}
\usepackage{algorithm}
\usepackage{algpseudocode}
\usepackage{multirow}
\usepackage{lipsum}
\usepackage{graphicx}
\usepackage{graphics}
\usepackage{color}
\usepackage[T1]{fontenc}
\usepackage{txfonts}
\usepackage{mathptmx}
\usepackage{booktabs}
\usepackage{textcomp}
\usepackage{CJK}
\usepackage{wrapfig}
\usepackage{courier}
\theoremstyle{definition}

\usepackage{colortbl}
\usepackage{color,xcolor}

\newcommand{\nop}[1]{}

\title{ToReMi: Topic-Aware Data Reweighting for Dynamic \\Pre-Training Data Selection}

\author{
Xiaoxuan Zhu\textsuperscript{\rm $\spadesuit$}\thanks{Equal Contribution},
Zhouhong Gu\textsuperscript{\rm $\spadesuit$*},
Baiqian Wu\textsuperscript{\rm $\spadesuit$},
Suhang Zheng\textsuperscript{\rm $\heartsuit$},
Tao Wang\textsuperscript{\rm $\heartsuit$},
Tianyu Li\textsuperscript{\rm $\heartsuit$},\\
\textbf{Hongwei Feng\textsuperscript{\rm $\spadesuit$}\thanks{Corresponding authors.}\ ,
Yanghua Xiao\textsuperscript{\rm $\spadesuit$$\dagger$}}\\
\textsuperscript{\rm $\spadesuit$}Shanghai Key Laboratory of Data Science, School of Computer Science, Fudan University\\
\textsuperscript{\rm $\heartsuit$}Alibaba Group\\
\{xxzhu22, zhgu22\}@m.fudan.edu.cn\\
\{suhang.zhengsh, shayue.wt, qianchuan.lty\}@alibaba-inc.com \\
\{hwfeng, shawyh\}@fudan.edu.cn
}

\begin{document}
\begin{CJK}{UTF8}{gbsn}
\maketitle
\begin{abstract}

Pre-training large language models (LLMs) necessitates enormous diverse textual corpora, making effective data selection a key challenge for balancing computational resources and model performance. 
Current methodologies primarily emphasize data quality metrics and mixing proportions, yet they fail to adequately capture the underlying semantic connections between training samples and quality disparities within individual domains. 
We introduce ToReMi (Topic-based Reweighting for Model improvement), a novel two-stage framework that dynamically adjusts training sample weights according to their topical associations and observed learning patterns. 
Our comprehensive experiments reveal that ToReMi variants consistently achieve superior performance over conventional pre-training approaches, demonstrating accelerated perplexity reduction across multiple domains and enhanced capabilities on downstream evaluation tasks. 
Code is available at \url{https://github.com/zxx000728/ToReMi}.
\end{abstract}

\section{Introduction}
\begin{figure}[t]
\centering
\resizebox{\columnwidth}{!}{
\includegraphics[width=1\linewidth]{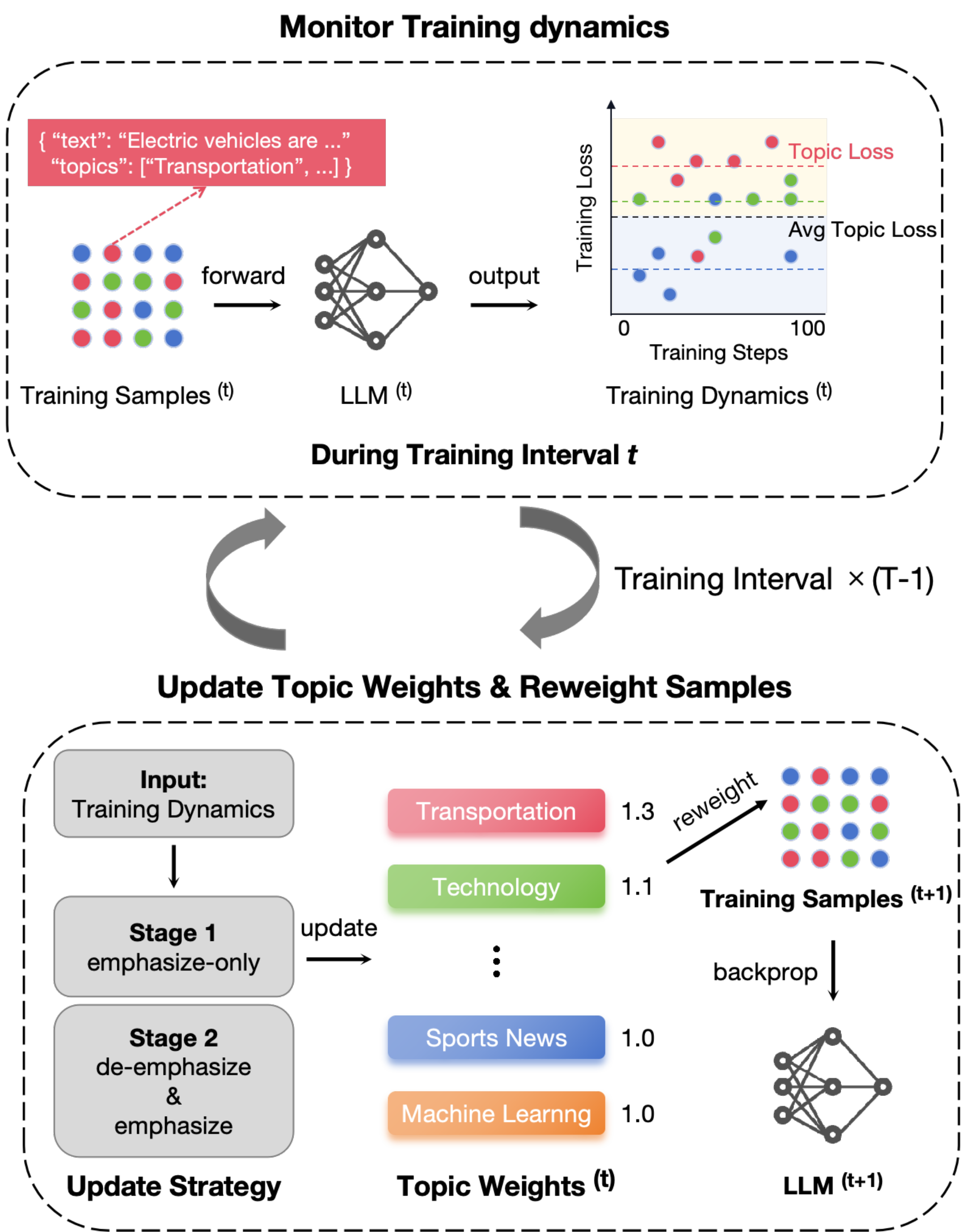}
}
\caption{
The framework of ToReMi, a two-stage, topic-based reweighting method for dynamic pre-training data selection and model improvement.
During each training interval, training samples are reweighted based on their topic labels and previous training dynamics.
}
\label{fig:framework}
\end{figure}

Large language models (LLMs) typically undergo pre-training on extensive corpora derived from heterogeneous sources of varying quality~\citep{gao2020pile, soldaini2024dolma, penedo2023refinedweb}. 
As model parameters and pre-training datasets continue to scale~\citep{kaplan2020scalinglawsneurallanguage, hoffmann2022trainingcomputeoptimallargelanguage}, the pre-training phase has emerged as the critical determinant of an LLM's foundational knowledge acquisition and reasoning capabilities~\citep{zhou2023limaalignment}. 
Consequently, systematic optimization of pre-training data constitutes a fundamental technical challenge in developing high-performance LLMs.

Current pre-training data optimization methodologies primarily address two complementary dimensions: 
quality assessment and distribution optimization. 
Both approaches aim to maximize the utility of pre-training data by prioritizing valuable content while mitigating potentially detrimental samples. 
Conventional pre-processing pipelines incorporate language identification, corpora filtration, deduplication, and noise reduction~\citep{soldaini2024dolma, penedo2023refinedweb, albalak2024survey}. 
Quality assessment mechanisms predominantly utilize rule-based heuristics and supervised classification models~\citep{raffel2023exploring, rae2022scaling, longpre2023pretrainers}, while distribution optimization refines corpus composition through calibrated domain ratio adjustments and strategic sampling techniques~\citep{xie2023doremi, du2022glam, soldaini2024dolma, thrush2024improvingpretrainingdatausing} to enhance model generalization capabilities.

Despite these advancements, constructing optimal pre-training datasets presents persistent challenges. 
Quality assessment approaches based on rules and classifiers remain inherently constrained by subjective annotation biases and limited training samples, effectively filtering only conspicuously low-quality content while failing to discern subtle quality variations~\citep{wenzek2019ccnet, xie2023data}. 
Similarly, current distribution optimization techniques employ relatively rudimentary methods, primarily validating effectiveness through proportional adjustments across topical or domain categories without adequately addressing intrinsic semantic relationships or dynamic training requirements~\citep{xie2023doremi, du2022glam}. 
These limitations collectively impede improvements in pre-training efficiency and model performance.

To address these limitations, we investigate a fundamental research question: 
How can pre-training dynamically prioritize high-quality data while accounting for both latent semantic relationships within the corpus and intra-domain quality variations?
We propose a two-stage \textbf{To}pic-based \textbf{Re}weighting framework for \textbf{M}odel \textbf{i}mprovement (\textbf{ToReMi}) in response to these challenges.
ToReMi's methodological innovation resides in its collective weight adjustment mechanism operating on topic categories. 
Rather than optimizing individual sample weights, it dynamically recalibrates entire topic categories based on the aggregate performance of constituent samples during training. 
The framework operates through two sequential phases: 
(1) During initial training, the system assigns elevated weights to challenging topic categories, prioritizing the learning of these hard samples; 
(2) Subsequently, the system progressively attenuates weights for underperforming topic categories (potentially containing higher noise concentrations) to minimize interference effects. 
Through this topic-level collective adjustment strategy, ToReMi optimizes pre-training data distribution without additional computational overhead while providing interpretable analysis of topic-specific training impact through weight trajectory feedback.

To rigorously evaluate ToReMi's efficacy, we conducted comprehensive experiments using the GPT-2 architecture~\citep{raffel2023exploring, rae2022scaling}. 
The experimental corpus comprised 2.6B tokens of curated Wikipedia content, semantically partitioned into 39 topics through large language model annotation. 
Experimental results demonstrate that ToReMi consistently outperforms both standard pre-training protocols and enhanced noise-resistant baselines in log perplexity evaluations on the Paloma corpus~\citep{gao2020pile}. 
In noise-injection experiments, ToReMi achieved 1.9\% average performance improvements on GLUE benchmarks compared to standard pre-training approaches~\citep{longpre2023pretrainers}. 
Further robustness analysis confirms that ToReMi maintains performance advantages across varied hyperparameter configurations, demonstrating methodological stability and adaptability.
\section{Related Work}
\subsection{Pretraining Data Filtering}
Pre-training data filtering has been extensively studied to enhance model performance and training efficiency~\citep{liu2024datasets, albalak2024survey}.
Common steps typically include language filtering~\citep{bigscience, chowdhery2022palm}, quality filtering~\citep{raffel2023exploring, rae2022scaling}, content filtering~\citep{xu2021detoxifying, longpre2023pretrainers}, and deduplication~\citep{hernandez2022scaling, lee2022deduplicating}.
Filtering methods generally fall into two categories: heuristic-based and classifier-based.
Heuristic methods use manually designed rules derived from corpus characteristics~\citep{penedo2023refinedweb, bigscience, raffel2023exploring}, while classifier-based methods train classifiers to assign quality scores~\citep{brown2020language, gao2020pile, xie2023data}.
Deduplication, on the other hand, typically uses hash-based techniques~\citep{bloom1970space, wenzek2019ccnet} for exact matching and model-based methods~\citep{abbas2023semdedup} for approximate matching.
While these approaches significantly improve corpus quality, their static nature hinders dynamic adjustments during training, making them prone to discarding valuable data~\citep{muennighoff2023scaling} and introducing biases~\citep{gururangan-etal-2022-whose, longpre2023pretrainers, dodge2021documenting}.

\subsection{Pretraining Data Mixing}
Pre-training datasets are often sourced from diverse domains, making effective data mixing strategies essential for maximizing their utility.
Fixed data mixing proportions, commonly used in practice~\citep{gao2020pile, rae2022scaling, touvron2023llama, soldaini2024dolma}, often rely on intuition and heuristics, such as upsampling high-quality domains like academic texts.
To automate this process, ~\citep{xie2023doremi} trains a reference model to guide proxy model training by minimizing worst-case excess loss, while ~\citep{fan2024doge} learns domain weights that maximize proxy model generalization to target domains.
However, the static nature of these methods hinders their adaptability to evolving training dynamics, while the need to train multiple models further reduces their efficiency.
To address these limitations, online data mixing strategies have been proposed.
ODM~\citep{albalak2023efficientonlinedatamixing} dynamically adjusts domain weights at each iteration to prioritize domains that reduce perplexity most effectively.
Skill-it~\citep{chen2023skillit} accelerates skill acquisition by leveraging the inherent order of prerequisite skills in the data.
Additionally, ~\citep{ye2024datamixinglaws} introduces data mixing laws to predict model performance for different data mixtures.
While these methods focus on inter-domain data mixing, intra-domain mixing of diverse data characteristics remains underexplored.

\section{Topic-Based Reweighting for Model Improvement (ToReMi)}
In this section, we introduce ToReMi (Figure~\ref{fig:framework}), a two-stage topic-based reweighting framework for dynamic pre-training data selection and model improvement, which adjusts sample weights based on their topic labels and model's training dynamics.

\subsection{Preliminary}
Training dynamics refer to statistical and performance metrics monitored throughout the model's training process, where high loss or prediction uncertainty is often used to identify challenging or noisy samples~\citep{thakkar2023selfinfluence, jiang2019acceleratingdeeplearningfocusing, swayamdipta}.
In this work, we track training loss to guide dynamic data reweighting and selection.
In specific, pre-training dataset ${\mathcal{D}}$ consists of ${\mathcal{N}}$ samples ${\{x_1, x_2, \ldots, x_N\}}$, where ${x_i}$ = \{text, ${\mathcal{L}_i}$\} and ${\mathcal{L}_i = \{\ell_1, \ell_2, \ldots\}}$ denotes topic labels assigned to sample ${x_i}$.
Let ${\mathcal{L} = \bigcup_{i=1}^{N} \mathcal{L}_i}$ denote the total set of all unique topic labels in the dataset.
For each topic label ${\ell_i \in \mathcal{L}}$, an associated weight ${w_{\ell_i}}$ is assigned, and initially, all weights are uniformly set to 1.

In LLM pre-training, for each sample $x_i$ with ground truth ${y_i}$, the training sample loss ${L(x_i)}$ is computed using the cross-entropy loss between the model's predicted probability distribution and the target ground truth labels, which is calculated as: 
\begin{equation}
    L(x_i) = -\frac{1}{T} \sum_{t=1}^{T} \log P(y_t \mid x_i, \theta)
\end{equation}
where $T$ is the sequence length of ${x_i}$.

For a specific label $\ell \in \mathcal{L}$, the training label loss $L_\ell$ is defined as the average loss over all samples containing $\ell$, which is calculated as:
\begin{equation}
    L_\ell = \frac{1}{|\mathcal{D}_{\ell}|} \sum_{x_i \in \mathcal{D}_{\ell}} L(x_i)
\end{equation}
where $\mathcal{D}_{\ell} = \{x_i \in \mathcal{D} : \ell \in \mathcal{L}_i\}$ is the subset of samples tagged with $\ell$.
The average label loss $L_\mathcal{L}$ is:
\begin{equation}
    L_\mathcal{L} = \frac{1}{|\mathcal{L}|} \sum_{\ell \in \mathcal{L}} L_{\ell}
\end{equation}

\subsection{ToReMi: Topic-Based Reweighting}
As a two-stage topic-based reweighting framework, ToReMi aims to prioritize high-quality and impactful data while minimizing the influence of noisy or less relevant data.
Reweighting is an effective approach for online data selection, as it dynamically adjusts the influence of individual samples during training, offering nuanced control without the need to exclude data outright.
Prior work~\citep{thakkar2023selfinfluence} computes the squared norm of a sample's gradient, showing that in the early stage of training, samples with higher scores are key contributors to the model's learning, while in later stage, such samples are more likely to represent noise or out-of-domain data.
Since samples with higher training loss generally produce larger gradients, ToReMi simplifies the process by monitoring training loss directly.

In the first stage, ToReMi focuses on samples with high training loss, prioritizing their learning to help the model efficiently acquire diverse and foundational knowledge.
To incorporate topic-level associations, sample weights are adjusted based on their relative topic weights.
Specifically, the entire training process is divided into multiple fixed training intervals ${\{t_1, t_2, \ldots, t_T\}}$.
Over a fixed training interval ${t}$, for each topic $\ell$ trained during ${t}$, we compute the training label loss ${L_\ell^{(t)}}$ for the topic and the average label loss ${L_\mathcal{L}^{(t)}}$ across all topics within the interval.
In the subsequent interval $t+1$, the sample loss is adjusted using the weight:
\begin{equation}
    w_{\ell}^{(t)} = \begin{cases}
\min(w_{\ell}^{(t-1)} + \alpha \cdot \Delta L_{\ell}^{(t)}, \beta) & \text{if } L_{\ell}^{(t)} > L_\mathcal{L}^{(t)} \\
1 & \text{otherwise}
\end{cases}
\end{equation}
where $\Delta L_{\ell}^{(t)} = L_{\ell}^{(t)} - L_\mathcal{L}^{(t)}$ is the difference between the topic's loss and the average label loss.
$\alpha$ is a scaling factor controlling the adjustment magnitude.
$\beta$ is the upper limit for label weights, preventing excessive upweighting and maintaining training stability.
The weighted sample loss is calculated by:
\begin{equation}
    {L_{w}^{(t+1)}(x_i)} = \min(\prod w_{\ell}^{(t)}, \beta) \cdot L^{(t+1)}(x_i), \quad \ell \in \mathcal{L}_i
\end{equation}
The weighted loss is then utilized for backpropagation, enabling the model to dynamically adapt its training focus.

In the second stage, the focus transitions to minimizing the impact of noisy data while further prioritizing high-quality samples.
The label weights are adjusted as follows:
\begin{equation}
    w_{\ell}^{(t)} = \begin{cases}
\max(w_{\ell}^{(t-1)} - \alpha \cdot \Delta L_{\ell}^{(t)}, \gamma) & \text{if } L_{\ell}^{(t)} > L_\mathcal{L}^{(t)} \\
\min(w_{\ell}^{(t-1)} + \alpha \cdot \Delta L_{\ell}^{(t)}, \beta) & \text{otherwise}
\end{cases}
\end{equation}
where $\gamma$ is the lower limit for label weights to ensure sufficient representation of all labels.
Then, the weighted sample loss is calculated as described in the first stage and utilized in backpropagation to guide the training process.
The complete algorithm is presented in Algorithm~\ref{alg:toremi}.

\begin{algorithm*}
\caption{Topic-Based Reweighting Framework for Model Improvement (ToReMi)}\label{alg:toremi}
\begin{algorithmic}[1]
\State \textbf{Input:} Training dataset $\mathcal{D}$ with samples $\{x_1, x_2, \ldots, x_N\}$, associated topic labels $\mathcal{L} = \{\ell_1, \ell_2, \ldots\, \ell_L\}$, label weights $\{w_{\ell_1}, w_{\ell_2}, \ldots, w_{\ell_L}\}$, training intervals $\{t_1, t_2, \ldots, t_T\}$, scaling factor $\alpha$, upper limit $\beta$, lower limit $\gamma$.
Initialize $w_\ell = 1$ for all $\ell \in \mathcal{L}$.
\For{$t = 1, 2, \ldots, T-1$}
    \State Compute $L_\ell^{(t)}$ and $L_\mathcal{L}^{(t)}$ for all $\ell \in \mathcal{L}^{(t)}$.
    \For{each $\ell \in \mathcal{L}^{(t)}$}
        \If{Stage 1}
            \State $w_\ell^{(t)} \gets 
                \begin{cases}
                    \min(w_\ell^{(t-1)} + \alpha \cdot \Delta L_\ell^{(t)}, \beta), & \text{if } L_\ell^{(t)} > L_\mathcal{L}^{(t)} \\
                    1, & \text{otherwise}
                \end{cases}$
        \ElsIf{Stage 2}
            \State $w_\ell^{(t)} \gets 
                \begin{cases}
                    \max(w_\ell^{(t-1)} - \alpha \cdot \Delta L_\ell^{(t)}, \gamma), & \text{if } L_\ell^{(t)} > L_\mathcal{L}^{(t)} \\
                    \min(w_\ell^{(t-1)} + \alpha \cdot \Delta L_\ell^{(t)}, \beta), & \text{otherwise}
                \end{cases}$
        \EndIf
    \EndFor
    \For{each sample $x_i \in \mathcal{D}^{(t+1)}$}
        \State Compute $L_w^{(t+1)}(x_i) \gets \min(\prod_{\ell \in \mathcal{L}_i} w_\ell^{(t)}, \beta) \cdot L^{(t+1)}(x_i)$.
    \EndFor
    \State Perform backpropagation using $L_w^{(t+1)}(x_i)$ to update model parameters.
\EndFor
\end{algorithmic}
\end{algorithm*}
\section{Topic Annotation}
Pre-training datasets are vast and encompass a wide range of topics and domains.
However, the scarcity of datasets with predefined topic labels makes it difficult to directly leverage labeled data for effective training.
Thus, we propose two methods for annotating topic labels to each sample within general pre-training corpora.

Given the growing volume of data and the computational costs, clustering algorithms are first applied to group similar samples based on their semantic features.
After forming the clusters, the generative capabilities of LLMs are utilized to assign meaningful topic labels.
This process involves extracting representative keywords from each cluster, which are then used to generate topic labels through LLMs. 
Specifically, there are two strategies: one where the LLM generates abstract and customized labels directly from the keywords, and another where it selectes the most relevant labels from a predefined taxonomy of topics.
The first strategy, \textit{Cluster\&Generate}, enables the creation of customized topic labels, which offers flexibility and makes it particularly useful for datasets that do not align with existing classification systems.
In contrast, the second strategy, \textit{Cluster\&Select}, maps clusters to an existing taxonomy, ensuring consistency and standardization across diverse datasets.
\section{Experiments}
\subsection{Experimental Setup}
\paragraph{Dataset and Model}
The pre-training dataset is sampled from Dolma-v1\_5-sample~\citep{soldaini2024dolma}, a high-quality English-only dataset curated from a diverse range of sources.
Input sequences consist of 1024 consecutive tokens randomly sampled from the dataset.
A total of 30B tokens are selected for pre-training the GPT-2~\citep{radford2019language} series models from scratch.
Tab.~\ref{tab:models} presents the various model parameter sizes and corresponding training token counts.
This setup follows the Chinchilla-optimal scaling law \citep{hoffmann2022trainingcomputeoptimallargelanguage}, which recommends training tokens to be 20 times the number of model parameters for different model sizes.
Due to computational constraints, we focus on experiments with the 124M parameter model in this work, with larger model experiments planned for future versions.

\begin{table}[ht]
    \centering
    \resizebox{\columnwidth}{!}{
        \begin{tabular}{lcc}
            \toprule
            \textbf{Model Name} & \textbf{\#Parameter} & \textbf{\#Training Tokens} \\
            \midrule
            GPT-2~\footnote{https://huggingface.co/openai-community/gpt2} & 124M & 2.6B \\
            GPT-2 Medium~\footnote{https://huggingface.co/openai-community/gpt2-medium} & 355M & 7.2B \\
            GPT-2 Large~\footnote{https://huggingface.co/openai-community/gpt2-large} & 774M & 15.6B \\
            GPT-2 XL~\footnote{https://huggingface.co/openai-community/gpt2-xl} & 1.5B & 30B \\
            \bottomrule  
        \end{tabular}
    }
    \caption{The model parameter sizes and the number of training tokens.}
    \label{tab:models}
\end{table}

\begin{figure}[ht]
\centering
\resizebox{\columnwidth}{!}{
\includegraphics[width=1\linewidth]{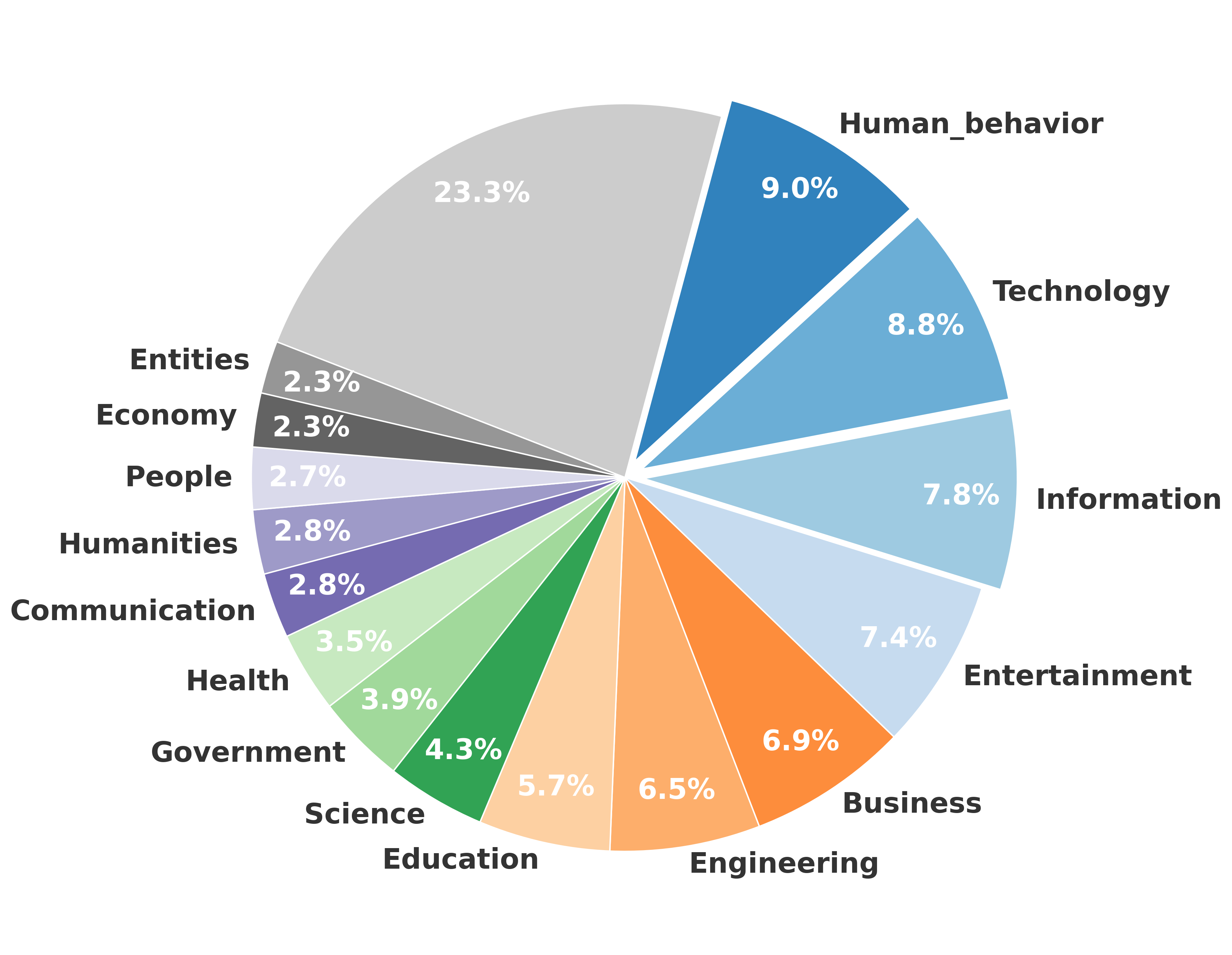}
}
\caption{
Topic distribution of the 30B-token dataset organized by Wikipedia taxonomy.
}
\label{fig:topic_distribution}
\end{figure}

\paragraph{Topic Annotaion Details}
For topic annotation, K-means clustering is first applied to group samples based on their embeddings generated by the BGE-M3 model~\citep{bge-m3}. 
Then, 100 representative keywords per cluster are extracted using TF-IDF. 
These keywords serve as input for Llama3-70B~\citep{llama3modelcard}, which is utilized to either generate topic labels directly or select the most relevant labels from Wikipedia's main topic classifications~\footnote{https://en.wikipedia.org/wiki/Category:\\Main\_topic\_classifications}.
The prompts employed for this purpose are detailed in Fig.~\ref{fig:prompt_generate} and Fig.~\ref{fig:prompt_select}. 
Fig.~\ref{fig:topic_distribution} presents the topic distribution of the entire 30B-token dataset as categorized according to the Wikipedia taxonomy.

\paragraph{Baselines}
We compare our two-stage ToReMi framework against two baseline approaches: standard pre-training (referred to as \textit{Standard}) and a partial implementation that applies only Stage 1 of our framework (denoted as \textit{ToReMi+Stage1}). 
The latter approach consistently prioritizes high-loss samples throughout training, similar to the strategy employed in Focal Loss \citep{lin2018focallossdenseobject}, which aims to enhance the model's capacity to learn from challenging samples.

\paragraph{Pre-training Settings}
During pre-training, training dynamics are monitored at intervals of $t=100$ steps.
The weight adjustment scaling factor $\alpha$ is configured with a default value of 1.0, while the upper and lower limits $\beta$ and $\gamma$ are set to 5.0 and 0.1 respectively.
The reweighting mechanism transitions from Stage 1 to Stage 2 after completing 4,000 training steps.

\paragraph{Evaluation Settings}
The evaluation of ToReMi encompasses two primary aspects.
For language modeling capabilities, we measure perplexity on the Paloma dataset~\citep{PalomaAB} to evaluate how well the model fits to language distributions in diverse domains.
Specifically, Paloma contains data collected from 12 distinct sources, all of which are held out from the pre-training corpus.
For downstream task performance, the GLUE benchmark~\citep{wang2019glue} (i.e., CoLA, SST-2, MRPC, QQP, STS-B, MNLI, QNLI, RTE, and WNLI) is utilized, which covers various dimensions of language understanding from grammaticality judgment to natural language inference.
Additionally, we also evaluate on PIQA~\citep{Bisk2020} for physical commonsense reasoning and SciQ~\citep{SciQ} for scientific knowledge assessment.
Both tasks are selected according to the Pythia scaling experiment~\citep{biderman2023pythiasuiteanalyzinglarge}, which demonstrates that models with approximately 160M parameters perform meaningfully above chance.

\subsection{Overall Performance}
\label{overall}

\begin{figure*}[t]
\centering
\resizebox{\textwidth}{!}{\includegraphics[width=1\linewidth]{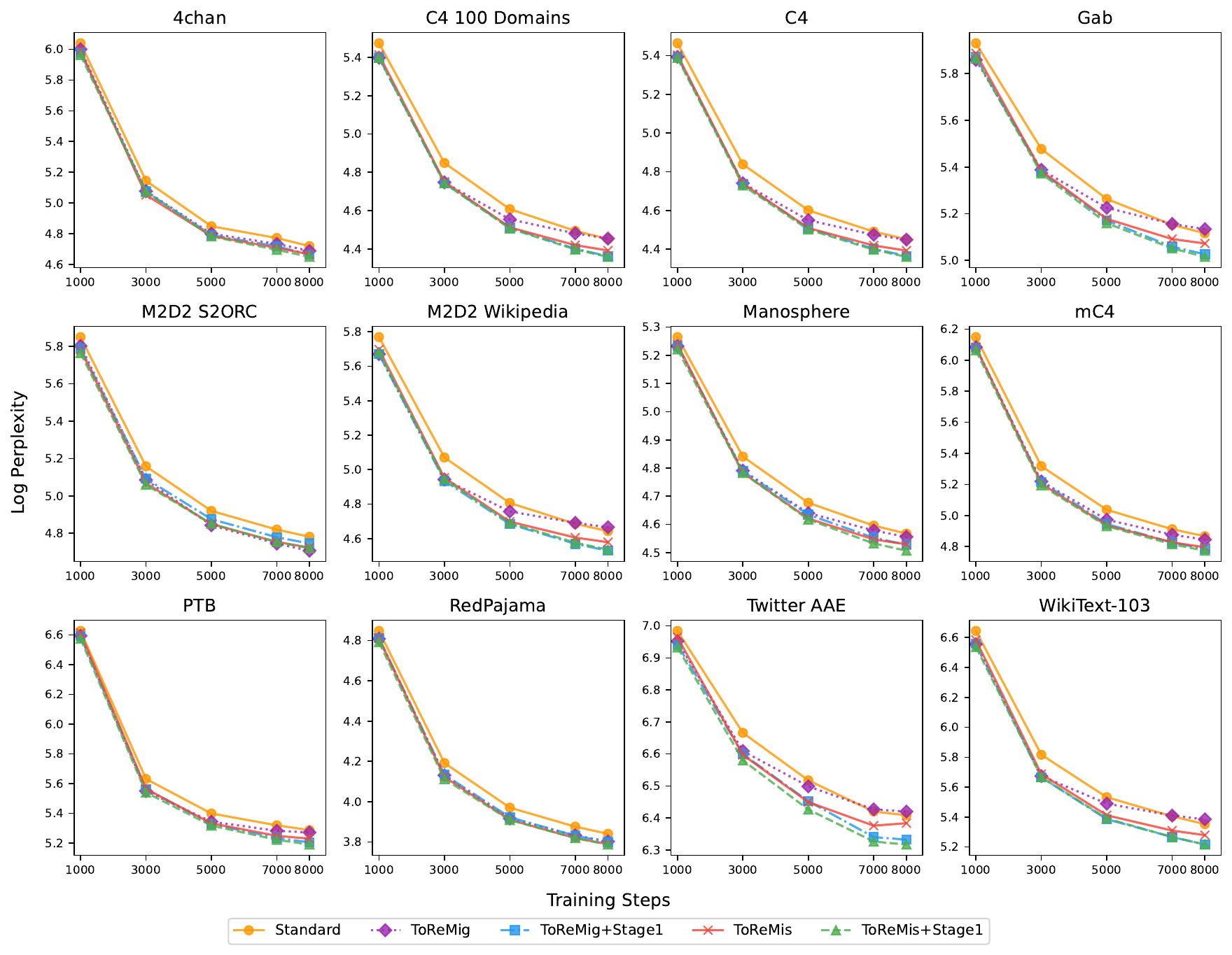}}
\caption{The log perplexity for different methods on the Paloma test dataset across 12 domains. ToReMig refers to ToReMi with directly generated topic labels, and ToReMis refers to ToReMi with topic labels selected from Wikipedia taxonomy.}
\label{fig:perplexity}
\end{figure*}

The experimental results are presented in Fig.~\ref{fig:perplexity} and Tab.~\ref{tab:overall}.
As is shown in Fig~\ref{fig:perplexity}, all ToReMi variants consistently reduce perplexity more rapidly than the standard method in all domains, particularly during steps 1000-5000, indicating faster convergence with the topic-based reweighting mechanism.
By final training steps, ToReMi achieves lower perplexity scores than the standard method in most datasets, indicating better overall language modeling capability.

Furthermore, the first section of Tab.~\ref{tab:overall} reveals that ToReMi's impact on downstream tasks is task-dependent.
For example, $ToReMi_S+Stage1$ improves by 5.78\% over standard method on CoLA, and all ToReMi variants show consistent gains on SST-2.
However, standard method outperforms on tasks like STS-B and RTE.
This pattern indicates that topic-based reweighting has varying effects on different linguistic capabilities.
ToReMi excels in tasks requiring broad linguistic patterns across diverse topics, strengthening foundational representations for syntactic understanding and sentiment analysis. 
Conversely, specialized reasoning tasks benefit from exposure to difficult examples that may be underrepresented after reweighting. 
The downweighting mechanism, while reducing noise, potentially limits exposure to challenging but informative instances needed for complex reasoning and domain-specific tasks.

\begin{table*}[ht]
    \centering
    \resizebox{\textwidth}{!}{
        \begin{tabular}{lcccccccccccc}
            \toprule
            \textbf{Method} & \textbf{CoLA} & \textbf{SST-2} & \textbf{MRPC} & \textbf{STS-B} & \textbf{QQP} & \textbf{MNLI} & \textbf{QNLI} & \textbf{RTE} & \textbf{WNLI} & \textbf{SciQ} & \textbf{PIQA} & \textbf{Overall}\\
            \midrule
            \multicolumn{13}{l}{\textit{\textbf{Overall Performance of Pre-training the 124M GPT-2 Model Using Different Methods}}} \\
            \midrule
            Standard & 17.49 & 86.58 & 75.37 & \textbf{74.64} & 84.42 & 75.10 & 82.81 & \textbf{59.20} & \textbf{56.33} & \textbf{24.60} & 56.31 & \textbf{62.99} \\
            $ToReMi_{S}+Stage1$ & \textbf{23.27} & \textbf{88.07} & \textbf{77.47} & 54.55 & \textbf{84.75} & 75.32 & 82.26 & 52.34 & 43.66 & 23.50 & \textbf{57.07} & 60.21 \\
            $ToReMi_{S}$ & 18.18 & 86.69 & 75.88 & 72.80 & 84.71 & \textbf{75.46} & 82.46 & 55.23 & 54.93 & 24.30 & 56.53 & 62.47 \\
            $ToReMi_{G}+Stage1$ & 15.93 & 87.27 & 76.34 & 72.48 & \textbf{84.75} & 75.24 & \textbf{82.96} & 57.04 & 43.66 & 23.50 & 56.91 & 61.46 \\
            $ToReMi_{G}$ & 16.84 & 87.61 & 76.36 & 73.24 & 84.72 & 75.26 & 82.04 & 54.15 & 42.25 & 23.70 & 56.31 & 61.13 \\
            \midrule
            \multicolumn{13}{l}{\textit{\textbf{Pre-training the 124M GPT-2 Model on Synthetic Noise Text}}} \\
            \midrule
            Standard & 17.79 & 86.35 & 74.40 & 71.18 & 84.09 & 75.08 & 81.84 & 48.01 & \textbf{54.93} & \textbf{24.60} & 55.88 & 61.29 \\
            $ToReMi_{S}+Stage1$ & \textbf{22.81} & \textbf{86.81} & 74.43 & 69.09 & \textbf{84.47} & 75.36 & 81.69 & \textbf{54.51} & 43.66 & 23.20 & \textbf{56.91} & 61.18 \\
            $ToReMi_{S}$ & 21.35 & 86.69 & \textbf{76.23} & \textbf{73.25} & 84.39 & \textbf{75.61} & \textbf{82.15} & 51.98 & 43.66 & 25.20 & 56.20 & \textbf{61.52} \\
            \midrule
            \multicolumn{13}{l}{\textit{\textbf{Effect of Stage Transition Point}}} \\
            \midrule
            $ToReMi_{S}+3000step$ & 20.79 & 86.81 & 75.48 & 68.71 & 84.12 & 75.04 & 81.73 & 52.70 & 43.66 & \textbf{27.40} & \textbf{56.80} & 61.20 \\
            $ToReMi_{S}+4000step$ & 21.35 & 86.69 & 76.23 & \textbf{73.25} & 84.39 & \textbf{75.61} & 82.15 & 51.98 & 43.66 & 25.20 & 56.20 & 61.52 \\
            $ToReMi_{S}+5000step$ & 13.01 & \textbf{87.50} & 74.68 & 72.78 & 84.37 & 75.46 & \textbf{82.75} & \textbf{58.12} & 38.02 & 24.00 & \textbf{56.80} & 60.68 \\
            $ToReMi_{S}+6000step$ & \textbf{22.62} & 86.35 & 75.24 & 68.72 & 84.52 & 75.24 & 82.20 & 49.45 & \textbf{53.52} & 27.00 & 56.31 & \textbf{61.92} \\
            $ToReMi_{S}+7000step$ & 20.13 & \textbf{87.50} & \textbf{77.06} & 70.30 & \textbf{84.54} & 75.57 & 82.39 & 54.51 & 42.25 & 25.40 & 56.69 & 61.49 \\
            \midrule
            \multicolumn{13}{l}{\textit{\textbf{Effect of Reweighting Bounds $(\gamma,\beta)$}}} \\
            \midrule
            $ToReMi_{S}+Stage1+(1.0,5.0)$ & 22.81 & 86.81 & 74.43 & 69.09 & 84.47 & 75.36 & 81.69 & 54.51 & 43.66 & 23.20 & \textbf{56.91} & 61.18 \\
            $ToReMi_{S}+Stage1+(1.0,10.0)$ & \textbf{24.11} & \textbf{87.72} & 76.79 & \textbf{74.62} & \textbf{84.72} & 75.29 & \textbf{83.39} & \textbf{59.20} & 36.62 & 23.90 & 56.26 & 62.06 \\
            $ToReMi_{S}+Stage1+(1.0,20.0)$ & 19.73 & 87.38 & 76.80 & 71.43 & 84.66 & \textbf{75.64} & 82.02 & 51.62 & 40.84 & 24.80 & 56.58 & 61.05 \\
            $ToReMi_{S}+(0.1,5.0)$ & 21.35 & 86.69 & 76.23 & 73.25 & 84.39 & 75.61 & 82.15 & 51.98 & 43.66 & 25.20 & 56.20 & 61.52 \\
            $ToReMi_{S}+(0.1,10.0)$ & 21.31 & 86.23 & \textbf{76.91} & 73.06 & 84.37 & 75.35 & 82.04 & 57.04 & \textbf{56.33} & \textbf{25.80} & 56.64 & \textbf{63.19} \\
            $ToReMi_{S}+(0.1,20.0)$ & 20.68 & 85.78 & 75.05 & 63.26 & 84.50 & 75.19 & 82.31 & 49.81 & \textbf{56.33} & 24.80 & 55.98 & 61.24 \\
            \bottomrule
        \end{tabular}
    }
    \caption{Model performance using different pre-training methods on downstream tasks. The table presents results for: (1) pre-training with normal data, (2) pre-training with synthetic noise data, (3) effect of various stage transition points, and (4) effect of different reweighting bounds.}
    \label{tab:overall}
\end{table*}

\subsection{Synthetic Experiment}
\label{synthetic}
To further evaluate the effectiveness of ToReMi in dynamically selecting high-quality data during pre-training, a synthetic experiment was conducted by injecting noise into samples associated with a specific topic label.
The \textit{Technology} label, which accounts for a significant proportion of the dataset and represents an important domain for evaluation, was selected for this purpose.
Noise was introduced by randomly shuffling all characters within each sample to simulate low-quality data.
For better reproducibility, ToReMi with the Wikipedia topic classification ($ToReMi_S$) was adopted for all subsequent experiments.

The results are presented in the second section of Tab.~\ref{tab:overall}.
Standard pre-training performs poorly on most metrics, indicating that noisy samples significantly impede model learning.
ToReMi with Stage1 achieves notable gains in CoLA (5.02\%) and RTE (6.5\%), demonstrating that prioritizing high-loss labels in early training enhances the model's linguistic understanding, strengthening its grasp of both grammatical structures and semantic relationships.
The complete two-stage ToReMi achieves the highest overall score (61.52) with substantial improvements on both MRPC and STS-B compared to the standard method and Stage1-only variant, highlighting how effectively its downweighting strategy mitigates the impact of noisy data.



\subsection{Ablation Experiment}
\label{ablation}
\paragraph{Effect of Stage Transition Point}
To investigate the impact of stage transition point between training phases in ToReMi, we conducted experiments by varying the step at which training switches from weighting (Stage 1) to de-weighting (Stage 2) on the noisy dataset introduced in Sec.~\ref{synthetic}.
While the default transition occurs at 4000 steps within a total of 8000 steps, additional experiments were conducted with the transition points at \{3000, 5000, 6000, 7000\} steps.
We primarily focused on delayed transitions, as entering Stage 2 prematurely before model convergence results in downweighting certain topics before adequate learning, decreasing pre-training efficiency.

Results presented in the third section of Tab.~\ref{tab:overall} indicate that transition timing significantly impacts model performance.
The 6000-step transition point achieved the highest overall score (61.92), effectively balancing the initial aggressive learning phase with the subsequent noise-reduction phase.
This point provides sufficient time for the model to learn important patterns while still allowing adequate time to downweight noisy samples.
In contrast, the 5000-step point produced the lowest performance with a significant drop in CoLA (13.01) despite achieving the highest RTE score (58.12), suggesting that delayed transitions may cause overfitting to noisy samples in certain tasks while benefiting others.
The non-linear relationship between transition point and model performance demonstrates that careful tuning of this hyperparameter is critical when applying ToReMi to different task settings.

\begin{figure}[ht]
\centering
\resizebox{\columnwidth}{!}{\includegraphics[width=1\linewidth]{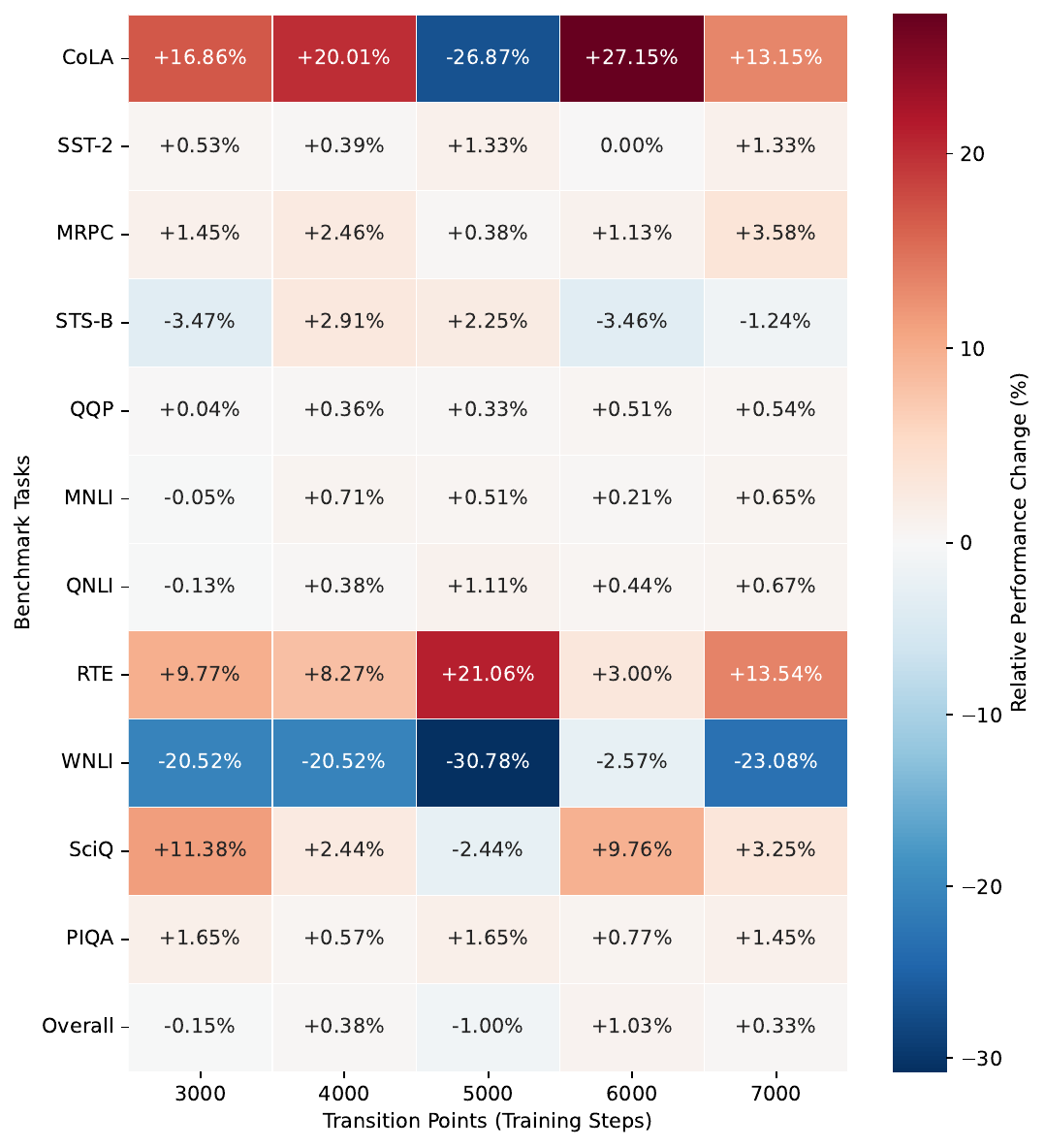}}
\caption{Performance difference between the standard method and ToReMi with various stage transition points. Red indicates performance improvement over the standard model, while blue indicates degradation.}
\label{fig:transition}
\end{figure}

Furthermore, Fig.~\ref{fig:transition} illustrates the performance difference between standard method and ToReMi with various stage transition points.
ToReMi outperforms standard method on most tasks regardless of transition point, with the exception of WNLI.
The consistent improvement on various tasks further validates the effectiveness and robustness of ToReMi.
The underperformance on WNLI is attributed to its unique characteristics as a natural language inference task with a small dataset (only 634 training examples).
WNLI requires understanding of complex pronoun resolution and discourse relationships, which are disproportionately affected by the topic-based reweighting mechanism.
The sample reweighting approach inadvertently downweights examples crucial for this particular task during Stage 2, indicating that specialized treatment is necessary for tasks heavily dependent on specific linguistic phenomena.

\paragraph{Effect of Reweighting Bounds}
To investigate the impact of reweighting bounds on model performance, we conducted experiments by varying the weight upper bound $\beta$ while maintaining a constant downweighting lower bound ($\gamma$ = 0.1).
The experiment prioritized upper bound analysis because excessively high weighting is susceptible to loss overexpansion for certain samples and introduces training instability, while the lower bound has comparatively smaller influence on overall performance.
Both the ToReMi+Stage1 variant and the complete ToReMi were evaluated with $\beta$ values of \{5.0, 10.0, 20.0\}.

The results are presented in the fourth section of Tab.~\ref{tab:overall}.
It is shown that moderate weight ($\beta$ = 10.0) produces optimal performance for both methods (62.06 for ToReMi+Stage1 and 63.19 for complete ToReMi), while further increasing $\beta$ to 20.0 causes degradation below even the $\beta$ = 5.0 configuration.
These findings indicate that increased weighting helps the model focus on challenging samples, though excessive upweighting leads to overfitting on particular topics and introduces instability in the training process.
Furthermore, when comparing the upweighting-only approach and the complete ToReMi at the same $\beta$ values, it is observed that the two-stage approach consistently outperforms the upweighting-only variant.
The performance gap is particularly pronounced at $\beta$ = 10.0, where the complete ToReMi achieved 1.13\% improvement.
The results highlight the importance of noise reduction during later training.
Initial upweighting enables the model to efficiently learn challenging topic-specific patterns, while subsequent downweighting reduces the influence of noisy samples, resulting in more robust performance on diverse tasks.

\begin{figure}[t]
\centering
\resizebox{\columnwidth}{!}{\includegraphics[width=1\linewidth]{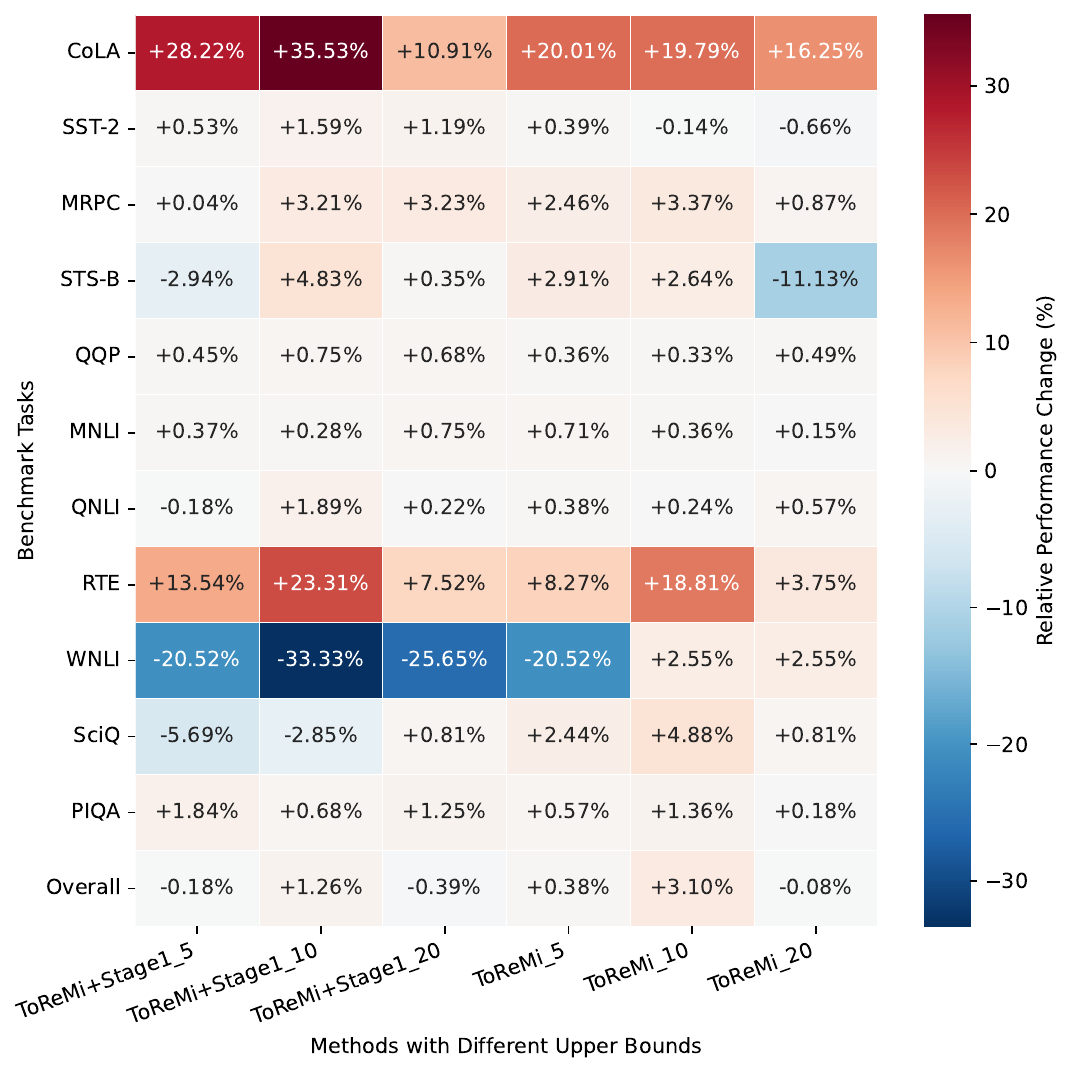}}
\caption{Performance difference between the standard method and ToReMi variants with various weight upper bounds. Red indicates performance improvement over the standard model, while blue indicates degradation.}
\label{fig:bounds}
\end{figure}

Fig.~\ref{fig:bounds} illustrates the performance difference between standard method and ToReMi variants with different weight upweighting upper bounds.
It is shown that both ToReMi and ToReMi+Stage1 outperform standard method on most tasks, demonstrating the effectiveness of our topic reweighting mechanism.
Notable improvements appear on CoLA and RTE, where ToReMi+Stage1 with $\beta$ = 10.0 achieves substantial gains of 35.53\% on CoLA and 23.31\% on RTE.
However, ToReMi+Stage1 also underperforms on WNLI, indicating that the sole upweighting potentially leads to overfitting on specific patterns, and the complete ToReMi (particularly with $\beta$ = 10.0) addresses this limitation through its downweighting strategy in later training phases.

\section{Conclusion}
In this paper, we introduced ToReMi, a novel two-stage data reweighting framework that dynamically adjusts sample weights based on corpus topics during pre-training to enable online data selection.
Experiments with GPT-2 pre-trained from scratch on the Dolma dataset demonstrate that ToReMi consistently outperforms standard methods, achieving faster perplexity reduction and lower final scores across domains.
ToReMi also shows particularly strong improvements on downstream tasks involving syntactic understanding and sentiment analysis, though benefits vary by tasks.
These findings establish topic-aware dynamic reweighting as a promising direction for improving both efficiency and effectiveness of language model pre-training.
Future work could further analyze which specific topic characteristics most benefit the pre-training process.

\bibliography{reference}

\newpage
\appendix
\section{Prompt for Topic Annotation}

\begin{figure}[ht]
\centering
\resizebox{\columnwidth}{!}{
\includegraphics[width=1\linewidth]{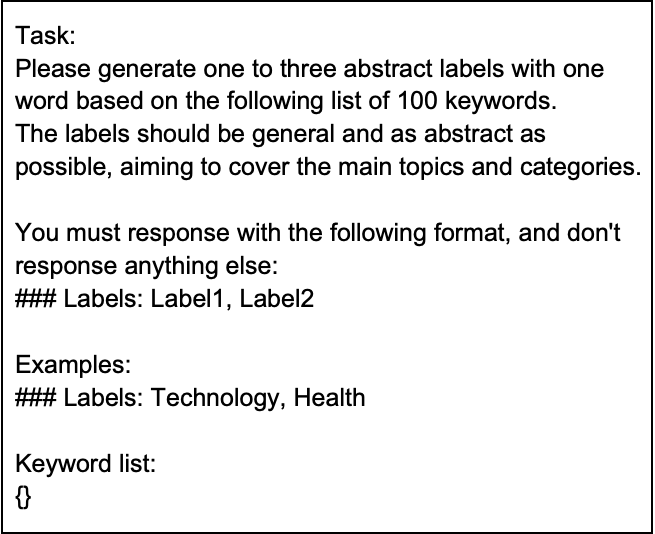}
}
\caption{
Prompt for generating topic labels for each sample using the provided extracted keywords.
}
\label{fig:prompt_generate}
\vspace{-3mm}
\end{figure}

\begin{figure}[ht]
\centering
\resizebox{\columnwidth}{!}{
\includegraphics[width=1\linewidth]{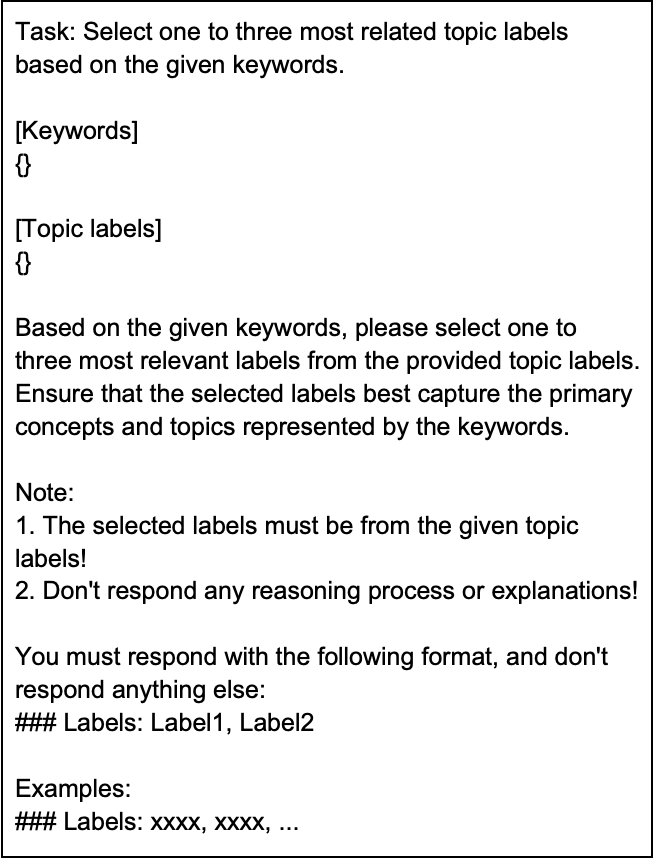}
}
\caption{
Prompt for assigning topic labels to each sample based on the provided Wikipedia taxonomy.
}
\label{fig:prompt_select}
\vspace{-3mm}
\end{figure}

\end{CJK}
\end{document}